\pdfoutput=1

\documentclass[11pt]{article}

\usepackage[final]{acl}

\usepackage{times}
\usepackage{latexsym}

\usepackage{enumitem}

\usepackage{xcolor,colortbl}

\usepackage[normalem]{ulem}
\usepackage{amsmath,amsfonts,amsthm,amssymb}

\newcommand{\com}[1]{}

\newcommand{\resolved}[1]{}

\definecolor{caribbeangreen}{rgb}{0.0, 0.8, 0.6}

\definecolor{lavender}{rgb}{0.58, 0.34, 0.92}

\newcommand{\nl}[1]{``\textit{#1}''}

\newcommand{\R}{\mathbb{R}}

\newtheorem{theorem}{Theorem}[section] 
\newtheorem{definition}[theorem]{Definition}

\usepackage[T1]{fontenc}

\usepackage[utf8]{inputenc}

\usepackage{microtype}

\usepackage{inconsolata}

\usepackage{graphicx}

\usepackage{booktabs}
\usepackage{siunitx}
\usepackage{multirow} 
\usepackage{makecell}
\usepackage{arydshln} 

\title{Confidence Improves Self-Consistency in LLMs}

\author{
\vspace{-0.2em} \\  
 \textbf{Amir Taubenfeld\textsuperscript{*12}},
 \textbf{Tom Sheffer\textsuperscript{*12}},
 \textbf{Eran Ofek\textsuperscript{1}},
 \textbf{Amir Feder\textsuperscript{13}},
 \textbf{Ariel Goldstein\textsuperscript{2}},
 \vspace{0.5em} \\  
 \textbf{Zorik Gekhman\textsuperscript{1}},
 \textbf{Gal Yona\textsuperscript{1}}
\\
\\
 \textsuperscript{1}Google Research,
 \textsuperscript{2}The Hebrew University of Jerusalem,
 \textsuperscript{3}Columbia University
\\
\vspace{-0.5em} \\  
    \{amirt,tomsheffer\}@google.com
}

\begin{document}
\maketitle

\def\thefootnote{*}\footnotetext{Equal contribution.}\def\thefootnote{\arabic{footnote}}

\begin{abstract}

Self-consistency decoding enhances LLMs' performance on reasoning tasks by sampling diverse reasoning paths and selecting the most frequent answer. However, it is computationally expensive, as sampling many of these (lengthy) paths is required to increase the chances that the correct answer emerges as the most frequent one.
To address this, we introduce \emph{Confidence-Informed Self-Consistency (CISC)}. CISC performs a \emph{weighted} majority vote based on confidence scores obtained directly from the model.
By prioritizing high-confidence paths, it can identify the correct answer with a significantly smaller sample size. 
When tested on nine models and four datasets, CISC outperforms self-consistency in nearly all configurations, reducing the required number of reasoning paths by over 40\% on average.
In addition, we introduce the notion of \emph{within-question confidence evaluation}, after showing that standard evaluation methods are poor predictors of success in distinguishing correct and incorrect answers to the same question. In fact, the most calibrated confidence method proved to be the least effective for CISC.
Lastly, beyond these practical implications, our results and analyses show that LLMs can effectively judge the correctness of their own outputs, contributing to the ongoing debate on this topic.
\end{abstract}

\begin{figure}[!ht]
    \centering
    \includegraphics[width=0.95\linewidth]{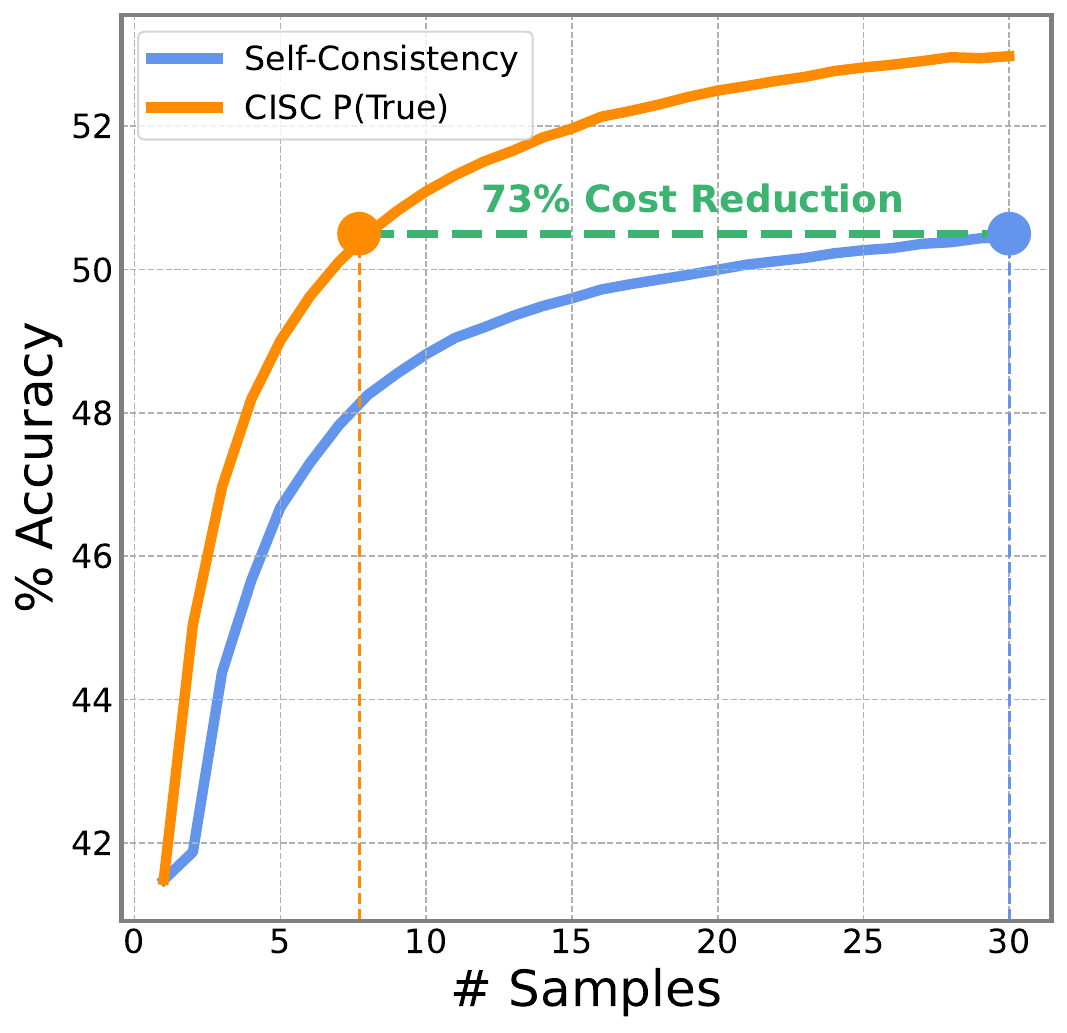}
    \caption{
    Accuracy as a function of the number of sampled responses for self-consistency vs CISC, using Gemma2-9B on the MATH dataset.
    CISC achieves higher overall accuracy while significantly reducing computational costs. With just 8 samples, it surpasses the performance of 30-sample self-consistency.}
    \label{fig:first-figure}
\end{figure}

\begin{figure*}[ht]
    \centering
    \includegraphics[width=\textwidth]{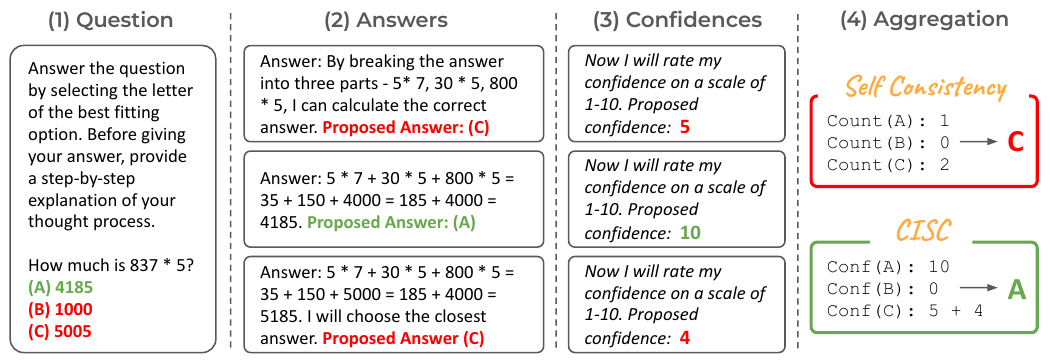}
    \caption{A simplified example comparing self-consistency vs CISC. (1) Given an input question, (2) both methods first sample multiple reasoning paths. (4, top) Self-consistency then simply selects the most frequent answer. Conversely, (3) CISC adds a self-assessment step, where a confidence score is assigned to each path (see \S\ref{sec:conf-methods} for more advanced methods). Then, (4, bottom) it selects the final answer via a weighted majority vote.}
    \label{fig:high-level}
\end{figure*}

\section{Introduction}
\label{sec:intro}

Modern large language models (LLMs) demonstrate strong reasoning capabilities \cite{bubeck2023sparks, guo2025deepseek}, 
driven in part by their capacity to generate a sequence of intermediate reasoning steps that lead them toward a final answer
\cite{wei2022chain, jaech2024openai}. 
Self-consistency \cite{wang2022self} is a popular decoding strategy that further improves LLMs' reasoning performance by sampling a diverse set of reasoning paths and selecting the most frequent answer as the final output. Despite its effectiveness, this approach is also computationally expensive, as it requires generating a large number of (long) reasoning paths to increase the chances that the correct answer emerges as the most frequent one.

Motivated by recent evidence that LLMs possess the ability to judge the correctness of their own outputs \cite{kadavath2022language, zhang2024small}, we hypothesize that self-consistency could be made significantly more efficient if the model could \emph{review} each generated reasoning path before selecting a final answer. We therefore introduce \textbf{Confidence-Informed Self-Consistency} (CISC), a lightweight extension of self-consistency. As illustrated in Figure \ref{fig:high-level}, CISC uses the model to generate a self-assessment score for each path and employs these scores in a weighted majority vote.

We conducted a comprehensive comparison of CISC and self-consistency, spanning nine LLMs of various sizes, four datasets covering a wide range of mathematical and commonsense reasoning tasks, and three popular methods for deriving self-assessment confidence scores from the model.
Our results demonstrate that CISC outperforms self-consistency in virtually all the examined configurations. Using the best-performing confidence estimation method, CISC achieves comparable performance to self-consistency while reducing the required number of reasoning paths by over 40\% on average (See Figure \ref{fig:first-figure} for an example).

Surprisingly, the most calibrated confidence method is actually the least useful for CISC. We offer a potential explanation:
existing confidence evaluation metrics measure the usefulness of confidence scores for comparing answers across different questions, while CISC requires distinguishing correct and incorrect answers for the same question.
To address this, we propose the Within-Question Discrimination (WQD) metric that specifically measures this ability, and demonstrate that it can predict the relative performance of CISC with different confidence methods.

Finally, we conduct a qualitative-analysis and find a significant agreement between model confidence scores and human assessments of the reasoning-paths' quality.  Specifically, responses identified by the model as low-confidence were also significantly more likely to be flagged by human evaluators as exhibiting signs of low-quality reasoning patterns.

To summarize, we contribute practical methods and foundational insights:
\begin{itemize}
    \item We propose CISC, a decoding strategy that can be used as a drop-in replacement to self-consistency, achieving comparable accuracy at a significantly lower computational cost.
    \item We introduce the concept of within-question confidence evaluation, after showing that standard evaluation methods are poor predictors of success in distinguishing correct and incorrect answers to the same question.
    \item We present empirical evidence supporting the idea that LLMs are capable of self-assessing their responses, contributing to the ongoing debate regarding this capability \cite{gero2023self, huang2023large, li2024confidence, stechly2024self}
    \item We open-source our implementation of CISC to facilitate further research\footnote{Our code is available at \url{https://github.com/google-research/google-research/tree/master/cisc}.}.
\end{itemize}

\section{Notations}

We consider an auto-regressive language model $M$ with parameters $\theta$. We use $p_\theta(\cdot \vert x)$ to denote $M$'s distribution over the next token given the provided context $x$. 
Given a question $q$ (e.g., \nl{Jane had 4 apples and ate half of her apples. How many apples she has now?}), we denote the model's response as $(\textbf{r}, \textbf{a})$,
where $\textbf{a}$ is the answer (e.g., \nl{2}) and $\textbf{r}$ is a \emph{reasoning path} (or chain-of-thought),  a sequence of logical steps supposedly leading up to this answer (e.g., \nl{If Jane ate half her apples, this means she ate 2 apples. 4 minus 2 is 2.}).

\section{Confidence-Informed Self-Consistency}
\label{sec:cisc}

In this section we present \textit{Confidence-Informed Self-Consistency} (CISC). 
When designing CISC, we hypothesized that it is possible to reduce self-consistency's computational costs by generating a \emph{confidence score} for each reasoning path, and performing a weighted majority vote.

As an intuitive example, consider a hypothetical setting where there exist only two possible answers, one correct and one incorrect. For a model that responds with the correct answer $60\%$ of the time, standard majority voting will require \emph{40 samples} to reach $90\%$ accuracy\footnote{Calculated using the binomial distribution. All the technical details are included in Appendix \ref{appendix:example}}. However, a weighted majority vote that weights correct answers twice as much as incorrect ones, will achieve 90\% accuracy with less than \emph{10 samples}. 

With this motivation in mind, we build on recent findings suggesting that LLMs are capable of judging the correctness of their own outputs \cite{kadavath2022language, tian2023just, zhang2024small}, and incorporate the model’s self-assessment of its reasoning paths into the final answer selection:

\begin{definition}[Confidence-Informed Self-Consistency]
\label{def:cisc}
Given a question $q$ and responses $\{(\textbf{r}_1, \textbf{a}_1), \dots, (\textbf{r}_m, \textbf{a}_m) \}$, CISC involves:

\begin{itemize}
    \item \textbf{Confidence Extraction}: A self-assessed confidence score $c_i\in\R$ is derived for each $(\textbf{r}_i, \textbf{a}_i)$.
    \item \textbf{Confidence Normalization}: The confidence scores are normalized
    using Softmax: $\tilde{c}_i = \frac{\exp\!\bigl(\frac{c_i}{T}\bigr)}{\sum_{j=1}^m \exp\!\bigl(\tfrac{c_j}{T}\bigr)}$, where $T$ is a tunable temperature hyper-parameter (see discussion below).
    \item \textbf{Aggregation}:  The final answer is selected using a confidence-weighted majority vote: $\hat{a}_{CISC} = \arg\max_a\sum_{i=1}^m \textbf{1}[\textbf{a}_i = a]\cdot \tilde{c}_i$. 
\end{itemize}
\end{definition}

The temperature parameter $T$ controls the relative importance of the answer frequency versus the confidence scores. Namely, as $T\to \infty$, the distribution of normalized confidence scores approaches the uniform distribution, and CISC collapses to vanilla self-consistency. Conversely, as $T\to 0$,  the softmax normalization approaches the hard maximum function, prioritizing the single response with the highest confidence and disregarding the overall frequency of answers. This may lead CISC to select a different answer than self-consistency (see Figure \ref{fig:high-level}). 

\section{Experimental Setup}
\label{sec:setup}

We compare CISC and self-consistency across a range of confidence extraction methods (\S\ref{sec:conf-methods}), reasoning tasks (\S\ref{sec:datasets}) and models (\S\ref{sec:models}).

\subsection{Confidence Extraction Methods}
\label{sec:conf-methods}

We use the following methods: 

\begin{itemize}[itemsep=1pt, topsep=2pt,leftmargin=*]
\item \textbf{Response Probability} \cite{wang2022self}: The confidence in a response $(\textbf{r}, \textbf{a})$  is taken to be the model's (length-normalized) probability of generating $(\textbf{r}, \textbf{a})=(x_1, \dots, x_n)$ given the question:$$p_\theta(\textbf{r}, \textbf{a}) = \left[\Pi_{i=1}^n p_\theta (x_i \vert x_1\dots x_{i-1}, q)\right]^\frac{1}{n}$$

\item \textbf{Verbal Confidence} \cite{lin2022teaching}: After sampling $(\textbf{r},\textbf{a})$ from the model, we prompt it to rate its confidence in its previously generated output. We implement two variants: (1) \textbf{Verbal Binary} instructs the model to output either 0 or 1, and (2) \textbf{Verbal 0-100} instructs the model to output a score on a scale of 0-100.

\item \textbf{P(True)} \citet{kadavath2022language}: We prompt the model to rate its confidence in $(\textbf{r},\textbf{a})$ in binary format (either 0 or 1), and compute the probability that the model assigns to the token $1$.

\end{itemize}

\paragraph{Efficient and Consistent Confidence Prompting.}

Our implementation of the prompt-based methods employs a \emph{two-step} prompting procedure (as depicted in Figure \ref{fig:high-level}).
Given a question prompt $q$, we first use the model to generate the reasoning chain and answer $(r,a)$. We then concatenate a confidence extraction prompt $e$ (e.g., \nl{Now I will rate my confidence...}), and continue the generation on $(q,r,a,e)$. This serves two important purposes. First, it ensures that when comparing self-consistency and CISC, the reasoning chains are identical. Second, the fact that the prefix $(q,r,a)$ remains unchanged after concatenating the confidence extraction prompt $e$ means it does not require reprocessing by the LLM. Consequently, the additional cost of the confidence extraction step consists only of encoding $\text{len}(e)\approx 20$ tokens and generating a single token. Since a single $(q,r,a)$ typically contains hundreds of tokens, the confidence extraction step adds only a negligible computational overhead to self-consistency. Further overhead reduction can be achieved through prompt optimization or by using the single-step procedure described in Appendix \ref{sec:appendix-prompting}. The precise prompts used and additional technical details are also provided in Appendix \ref{sec:appendix-prompting}.

\subsection{Datasets}
\label{sec:datasets}

We used four large reasoning benchmarks:\footnote{Other than the popular GSM8K, the other datasets were chosen as the three largest datasets in the Hugging Face Leaderboard \cite{leaderboard} (as of December 1st, 2024).}

\begin{itemize}[itemsep=1pt, topsep=2pt,leftmargin=*]
\item  \textbf{GSM8K} \cite{cobbe2021gsm8k}: A dataset of grade-school level math word problems. We evaluate on the entire validation set (1320 questions). 
\item \textbf{MATH} \cite{hendrycksmath2021}: A more challenging dataset of math word problems. We used the entire test set (5K questions). 
\item \textbf{MMLU-Pro} \cite{wang2024mmlu}: A more challenging version of the Multitask Language Understanding (MMLU)  benchmark, testing language models' general knowledge and reasoning abilities with a wide range of topics such as science and history. We randomly sampled 5K questions.
\item \textbf{Big-Bench-Hard} \cite{suzgun2022challenging}: A challenging selection of tasks from the big-bench benchmark \cite{srivastava2023beyond}, comprises a variety of reasoning tasks that pose challenges to LLMs, such as counting objects. We selected 20 out of 23 tasks (5,761 examples), eliminating three sub-tasks that required designated answer extraction methods.
\end{itemize}

\subsection{Models}
\label{sec:models}

We use nine instruction-tuned open-weights LLMs from 3 different families:

\begin{itemize}[itemsep=1pt, topsep=2pt,leftmargin=*]
\item \textbf{GEMMA2} \cite{team2024gemma}: A Google AI model family, including 2B, 9B, and 27B parameter models. 
\item \textbf{QWEN2.5} \cite{yang2024qwen2}: A model family from Alibaba AI, with 7 models ranging from 0.5B to 72B parameters. We selected three models: 3B, 14B, and 72B.
\item \textbf{Mistral} \cite{mistral}: We used three of the latest models available - Ministral-8B-Instruct-2410, Mistral-Small-Instruct-2409, mistralai/Mistral-Large-Instruct-2411 - with 8B, 22B, 123B parameters respectively.
\end{itemize}

\subsection{Metrics}
\label{sec:metrics}

We compare CISC against self-consistency using the following metrics:

\begin{itemize}[itemsep=1pt, topsep=2pt,leftmargin=*]

\item \textbf{\% Cost Reduction}: The percentage of computational cost saved by using CISC. We fix the compute budget for CISC (5 or 10 model responses) and measure the number of responses\footnote{If self-consistency failed to reach CISC's accuracy using up to 30 responses, we use a maximal value of 31 for this metric.} required for self-consistency to achieve equivalent accuracy:
$$100 \times \left(1 - \frac{\text{CISC budget}}{\text{\# Comparable SC responses}}\right)$$

\item \textbf{\% Accuracy Improvement}: The relative accuracy gain of CISC over self-consistency when both methods utilize the same number of responses per question: 
$$100 \times \left(\frac{\text{CISC Acc}}{\text{SC Acc}} - 1\right)$$
\end{itemize}

\subsection{Temperature Scaling}
\label{sec:temperature}

As discussed in \S\ref{sec:cisc}, CISC re-scales the confidence values using a softmax transformation, parameterized by a temperature $T > 0$. We tune the temperature separately for each model and confidence extraction method using a 10\% held-out set, aggregated across all four datasets (\S\ref{sec:datasets}). The fact that CISC only employees a single dataset-agnostic hyper-parameter, makes the tuning process light-weight and robust. More details and the optimal temperature values for each configuration are in appendix \ref{sec:appendix-temperature}.

\subsection{Bootstrap}
\label{sec:bootstrap}

To compute the performance of a decoding strategy $s$ (either self-consistency or a variant of CISC) with a sample budget of $b \in [1,...,30]$, we perform bootstrap sampling. We first sample $30$ different reasoning paths from the model. 
Next, we draw $n=500$ sets of $b$ paths for each question, apply $s$ to each set, and compute the accuracy per set. We then average the results across all bootstrap samples to obtain the final score.

\section{Main Results}

\label{sec:results}

\begin{table*}[!ht]
\centering
\begin{tabular}{lcccc}
\toprule
&  \multicolumn{2}{c}{Cost Reduction} & \multicolumn{2}{c}{Acc Improvement} \\
\cmidrule(lr){2-3} \cmidrule(l){4-5} 
Confidence Method & Budget 5 & Budget 10 & Budget 5 & Budget 10 \\
\midrule
Verbal Binary & 18\% \small{(6.1)} & 10\% \small{(11.1)} & 0.4\% & 0.2\% \\[4pt]
Verbal 1-100 & 22\% \small{(6.4)} & 30\% \small{(14.4)} & 0.8\% & 0.4\% \\[4pt]
Response Probability & 22\% \small{(6.5)} & 31\% \small{(14.6)} & 1.1\% & 0.8\% \\[4pt]
P(True) & \textbf{41\% \small{(8.4)}} & \textbf{46\% \small{(18.6)}} & \textbf{1.6\%} & \textbf{1.1\%} \\
\bottomrule
\end{tabular}
\caption{\textbf{CISC performance (macro-averaged over all datasets and models) per confidence method.}  
CISC performs better than standard self-consistency in terms of both efficiency gains and accuracy improvements across all confidence methods. Specifically, the \textbf{P-True} method achieves the best results. For instance, self-consistency must use 18.6 sampled responses on average to match the accuracy obtained by CISC using only 10 samples, representing a 46\% reduction in computational costs. }
\label{table:aggregated-results}
\end{table*}

This section demonstrates CISC's (\S\ref{def:cisc}) substantial performance advantage over self-consistency. We compare CISC, using fixed compute budgets of 5 and 10 responses per question, based on the metrics defined in \S\ref{sec:metrics}.

\paragraph{CISC outperforms self-consistency across virtually all models and datasets.}
Table \ref{table:aggregated-results} presents 
the \textit{Cost Reduction} and \textit{Accuracy Improvement} (see \S\ref{sec:metrics}) achieved by CISC with each confidence method. The results are macro-averaged across all models and datasets. CISC outperforms self-consistency with every confidence method.  

\begin{figure}[h]
    \centering
    \includegraphics[width=1\linewidth]{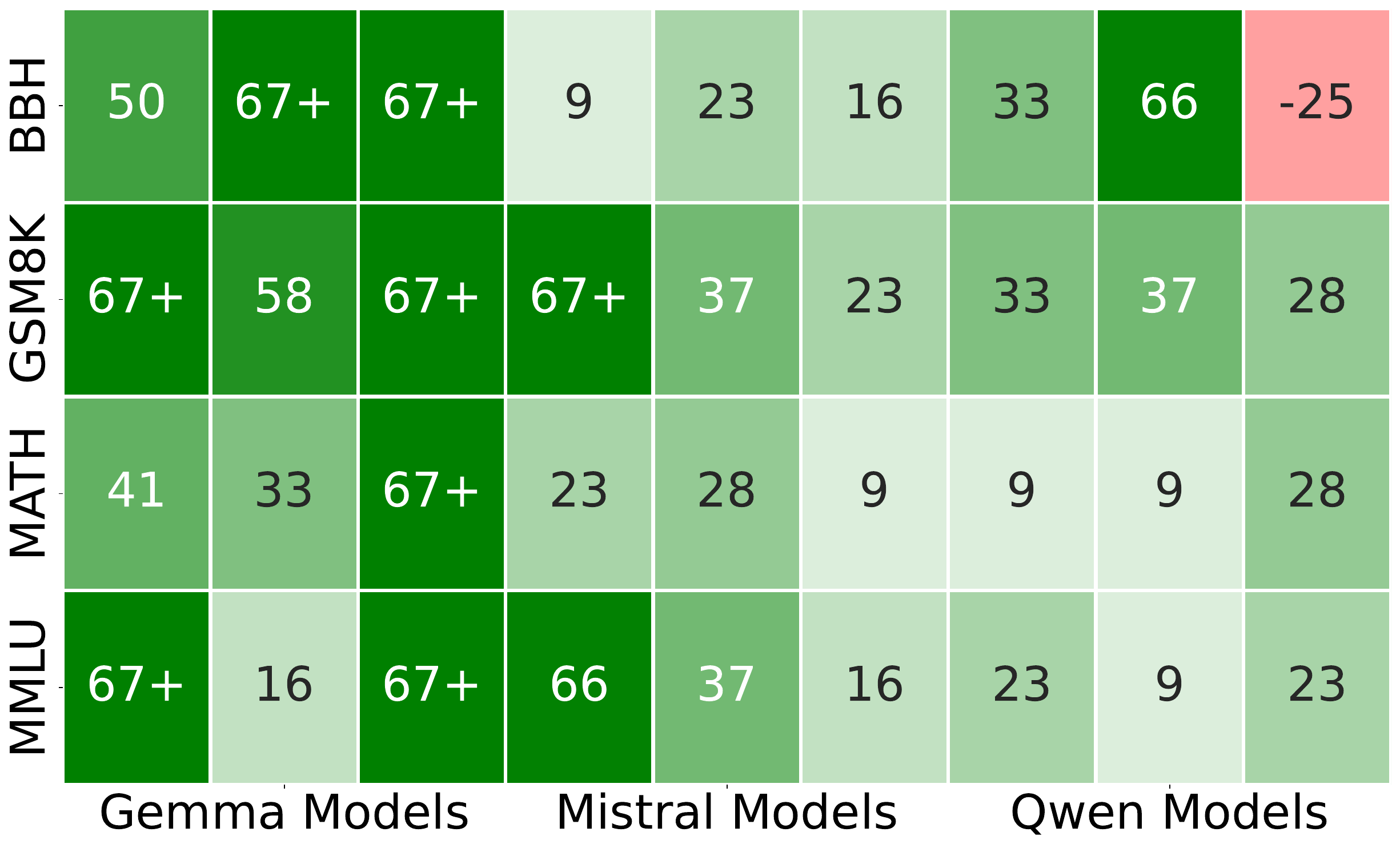}
    \caption{Results breakdown for CISC using the P(True) method and a budget of 10 responses per question. Each cell is annotated with the Cost Reduction (Percentage; \S\ref{sec:metrics}) of CISC compared to self-consistency. As can be seen, CISC improves performance across almost all model families and datasets. In many cases, even 30 samples are not enough for self-consistency to reach CISC performance, leading to Cost Reduction of over 67\%. }.
    \label{fig:intrinsic-binary-results}
\end{figure}

The \textit{P(True)} method yields the best results, achieving an average Cost Reduction of 41\% and 46\% with budgets of 5 and 10 responses, respectively. Figure \ref{fig:intrinsic-binary-results} presents a detailed breakdown of CISC's performance using \textit{P(True)} across all models and datasets. Notably, CISC is effective across nearly all configurations, with some exceeding 67\% cost reduction.

We provide additional results in Appendix \ref{sec:appendix-more-results}. In particular, Table \ref{tab:apendix-all-results} shows a per-dataset breakdown of Table \ref{table:aggregated-results}, and Table \ref{table:micro-results} shows the Accuracy Improvement metric micro-averaged across configurations, which enables the computation of confidence intervals. These demonstrate that the observed improvements of CISC (for each confidence method examined) are strongly statistically significant.

\paragraph{Confidence Normalization improves CISC's performance.} We drill down into the importance of the within-question confidence normalization step in CISC. In Table \ref{table:norm-table}, we compare CISC's performance with and without confidence normalization. We see that for every confidence method examined, CISC with normalization (softmax with a tunable temperature value) outperforms its unnormalized counterpart. However, as shown in Supplementary Table \ref{table:norm-table-ext}, normalization is effective only when using appropriate temperature hyper-parameters. Because different confidence extraction methods produce scores on different scales, their optimal temperatures vary considerably (values are provided in Supplementary Figure \ref{fig:temperatures-map}). For instance, the P(True) method yields confidence values with high similarity, thus requiring lower temperatures to distinguish between them.

\begin{table}[h]
\centering
\small
\begin{tabular}{lc}
\toprule
Confidence Method & \makecell{Cost Reduction @ 10} \\
\toprule
P(True) (w/o normalization) & 32\% \small{(14.8)} \\
P(True) (w/ normalization) & \textbf{46\% \small{(18.6)}} \\
\midrule
SP (w/o normalization) & 24\% \small{(13.1)} \\
SP (w/ normalization) & \textbf{31\% \small{(14.6)}} \\
\midrule
Verbal (w/o normalization) & 20\% \small{(12.5)} \\
Verbal (w/ normalization) & \textbf{30\% \small{(14.4)}} \\
\bottomrule
\end{tabular}
\caption{\textbf{CISC performance with and without confidence normalization} (bottom and top rows, respectively). We see that while  
CISC demonstrates substantial cost reductions even without normalization, employing normalization (Softmax and temperature scaling) significantly enhances performance, across all three confidence methods.}
\label{table:norm-table}
\end{table}
\section{Within-Question Confidence Evaluation}

\label{sec:pairwise}

\definecolor{lightgraycisc}{gray}{0.97}

\begin{table}[!ht]
\centering
\resizebox{\columnwidth}{!}{
\begin{tabular}{l c c c >{\columncolor{lightgraycisc}}c}
\toprule
\makecell{Confidence \\ Method} & ECE-t $\downarrow$ & Brier-t $\downarrow$ & \makecell{WQD $\uparrow$} & \makecell{CISC Cost\\ Reduction $\uparrow$} \\
\midrule
Verbal Binary & \textbf{0.005} & 0.187 & 52.2\%  & 10\% \\[4pt]
Verbal 0-100 & 0.046 & \textbf{0.173} & 56.1\%   & 30\% \\[4pt]
Response Prob. & 0.090 & 0.192 & 59.0\%    & 31\% \\[4pt]
P(True) & 0.030 & 0.182 & \textbf{62.3\%}        & \textbf{46\%} \\[4pt]
\bottomrule
\end{tabular}
}
\caption{\textbf{Comparison of different confidence extraction methods in terms of between-question and within-question confidence evaluation metrics}. 
We see that between-question metrics (ECE-t, Brier-t) are poor indicators of effective confidence extraction for CISC, while our novel WQD metric (\ref{def:wqd}) effectively predicts which confidence extraction method yields the best CISC performance.
}
\label{tab:confidence-methods}
\end{table}

Recent work demonstrated that verbal confidence methods significantly outperform P(True) in terms of \emph{calibration} \cite{tyen2023llms}, which is the de-facto approach to evaluate the quality of confidence measures. Yet, perhaps surprisingly, CISC is more effective with P(True) than with verbal confidence methods (Table \ref{table:aggregated-results}). In this section we settle these differences, and explain why well-calibrated confidence measures can still be less useful for CISC.

We argue that existing evaluation metrics, whether \emph{calibration} based \cite{kadavath2022language, tian2023just} or \emph{discrimination} based \cite{kuhn2023semantic, nguyen2024direct} examine the confidence behavior \emph{between} the input questions. However, for CISC to work well, we want the confidence scores to be able to distinguish correct and incorrect responses \emph{to the same question}. 

To gain an intuition for the difference between \emph{within-question}
and \emph{between-question} confidence evaluation, consider the following simple example. Imagine a model $M$ and a dataset with two types of questions: questions that $M$ finds \nl{easy} (e.g., answers correctly 95\% of the time) and questions that $M$ finds \nl{hard} (e.g., answers correctly 5\% of the time). Consider a confidence measure that assigns every answer to an \nl{easy} question a confidence of $0.95$ and every answer to a hard question a confidence of $0.05$. This confidence signal is useless for CISC, as it does not make any distinctions between answers to the same question. On the other hand, it scores well under existing metrics (e.g., it is perfectly calibrated).

The above thought experiment shows that the fact that well-calibrated confidence scores can be derived from a model does not necessarily imply the model possesses a capacity to self-assess its own responses. To isolate this specific ability, we design a metric that measures whether the confidence scores can distinguish correct and incorrect responses to the same question:

\begin{definition}[Within-Question Discrimination]
\label{def:wqd}
Given a dataset of questions, for each question $q$, denote the sampled responses by $R_q = \{(\textbf{r}_i, \textbf{a}_i)\}_{i=1}^m$, and let $R^+_q,R^-_q \subseteq R_q$ be the subsets of correct and incorrect responses respectively. We evaluate the Within-Question Discrimination (WQD) of a confidence method $c: (r,a) \mapsto \R$ as:
\begin{multline*}
    \text{WQD}(c) \equiv \\ 
    \frac{1}{N} \cdot \sum_{q} \sum_{(r, a)\in R^+_q} \sum_{(r', a')\in R^-_q} [c(r,a) > c(r',a')]
\end{multline*}
where $N = \sum_q |R_q^+| \cdot |R_q^-|$.
\end{definition}

That is, we compute the fraction of cases where the higher confidence response is indeed the correct response, out of pairs of responses \emph{to the same question} (exactly one of which is correct). In our work, we use $m = 30$ (as described in \S\ref{sec:bootstrap}).

To emphasize the importance of \emph{within-question evaluation}, we test if WQD is more predictive of CISC's success than standard \emph{between-question} confidence metrics. We compare each confidence method from \S\ref{sec:conf-methods} in terms of: (i) standard metrics, such as ECE \cite{pmlr-v70-guo17a} and Brier Score \cite{brier1950verification}, (ii) WQD, (iii) CISC performance at a budget of 10 samples. We follow previous work \cite{tyen2023llms} and report the standard metrics after applying temperature scaling \cite{ovadia2019can}, a technique that fits a single temperature parameter $T$ to the model's confidences to minimize the negative log-likelihood on the data. We use ECE-t and Brier-t to denote the scaled scores.

The results of this comparison, averaged across all datasets (\S\ref{sec:datasets}) and models (\S\ref{sec:models}), are summarized in Table \ref{tab:confidence-methods}.  
Indeed, we see that the verbal confidence methods obtain the best ECE-t and Brier-t scores while also achieving the worst performance in CISC. On the other hand, the WQD metric is able to perfectly predict the relative scores of each confidence method in CISC. This emphasizes the limitations of relying solely on traditional confidence evaluation methods for evaluating the models ability to self-assess its reasoning.

\begin{figure}[!h]
    \centering
    \includegraphics[width=1\linewidth]{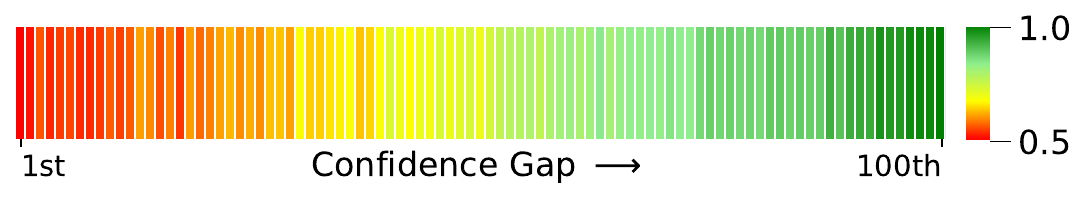}
    \caption{Within-Question Discrimination score (indicated by color) increases smoothly as a function of the confidence gap (percentiles, x-axis). Here we use the P(True) method, Gemma2-9B and the MATH dataset. 
    } 
\label{fig:pairwise-heatmap}
\end{figure}

The WQD metric prioritizes interpretability, focusing on the discrimination ability of the confidence scores irrespective of the relative magnitude of the confidence values $c(r,a)$ and $c(r', a')$. However, examining the relationship between WQD and the confidence gap $|c(r,a) - c(r',a')|$ offers additional insights. Figure \ref{fig:pairwise-heatmap} illustrates a near monotonic relationship: the within-question discrimination ability (indicated by color) smoothly increases with the confidence gap (x-axis).
These findings suggest a fine-grained self-assessment mechanism, where even small differences in confidence scores reflect significant variations in the probability of a response being correct

Taken together, our findings provide a compelling evidence that LLMs indeed posses an intrinsic ability to reassess their own responses.

\section{Qualitative Analysis}
\label{sec:qualitative}

In \S\ref{sec:results} we showed that CISC has clear 
performance advantages over standard self-consistency, and argued that this suggests LLMs are capable of self-assessing their confidence in responses to the same question. To facilitate a better understanding of this phenomenon, we asked human evaluators to identify indicators of \emph{low-quality model responses} (i.e., logical patterns that reduced the evaluators' confidence in the correctness of the LLM response). Our analysis revealed a strong correlation between the prevalence of these indicators and lower confidence scores assigned by the LLM.

\paragraph{Sampling Process.} We performed the analysis on MMLU-Pro (\S\ref{sec:datasets}), using three representative models, one from each model family.  

To reduce the evaluation burden we limited it to three LLM responses per question. We selected these triplets based on two criteria: (1) CISC and SC produced different results, where one method yielded a correct answer and the other did not, and (2) the final answers of the three responses were not all distinct, which would otherwise degenerate self-consistency's majority voting. 

Out of the remaining triplets, we randomly chose 45 for which SC was correct and 45 where SC was wrong. Then, for each triplet, we randomly took either the response with highest relative-confidence or the response with lowest relative-confidence. This ensured an equal number of low relative-confidence responses that were correct and incorrect, mitigating potential bias of answer correctness on our analysis. The process resulted in 90 responses for human evaluation.

\paragraph{Human Evaluation.} Two human evaluators (NLP Phd students), unaware of both the model's confidence scores and the ground truth labels, reviewed 90 samples. The evaluators' task was to identify logical patterns in the LLM reasoning-chain which reduce their confidence that the LLM has reached a correct answer; we call these patterns low-quality-indicators. Also, the evaluators were asked to briefly describe each identified pattern.

\paragraph{Results.} Our evaluation demonstrated a significant correlation in confidence assessments: 67\% of the samples assessed as relative-low confidence by the model were also judged to contain low-quality indicators by human evaluators, while only 33\% of the samples assessed as relative-high confidence by the model contained the human identified low-quality-indicators. This strong correlation suggests that LLMs are adept at assessing their own reasoning processes and identifying patterns that humans consider indicative of low quality. 

In addition, we categorized these low-quality indicators. Three primary categories emerged: (1) the LLM's final answer was not among the provided options; (2) the LLM deliberated between multiple options; and (3) the LLM omitted necessary calculations. Of these, only categories (1) and (3) showed a strong correlation with the LLM's low-confidence scores. Further details regarding these categories and their correlation statistics are available in the Appendix \ref{sec:appendix-qualitative}.
\section{Related Work}

\paragraph{Confidence signals for LLMs.} 
There is a long line of work on deriving confidence measures from LLMs. Popular approaches use
the agreement across multiple samples \cite{kuhn2023semantic, manakul2023selfcheckgpt, tian2023fine,lyu2024calibrating,gekhman-etal-2024-fine}, the model's internal representations \cite{azaria2023internal, burns2022discovering,orgad2025iclr} or directly prompting the model to verbalize its confidence \cite{tian2023just, kadavath2022language}.
All papers in this line of work focused on fact-seeking tasks, 
so confidence is typically derived based on the final answer alone. To the best of our knowledge, our work is the first to apply these approaches to scoring the entire reasoning path.

\paragraph{Reasoning verification.}
While learned verifiers have been demonstrated to significantly improve performance on math word problems \cite{cobbe2021training, lightman2023let, li2022making}, the ability of LLMs to perform \emph{self}-verification and \emph{self}-correction is still heavily contested, with some works providing positive evidence for such capabilities \cite{weng2022large, gero2023self, madaan2024self, liu2024large, li2024confidence} and others arguing that the gains can mostly be attributed to clever prompt design, unfair baselines, data contamination and using overly simple tasks \cite{tyen2023llms, valmeekam2023can, hong2023closer, huang2023large, stechly2024self, zhang2024small}. This work contributes to this ongoing discussion by presenting multiple lines of evidence supporting LLM self-verification. In particular, we demonstrate clear benefits from a simple confidence-based self-verification approach.

\paragraph{Improving self-consistency's efficiency. }

Numerous attempts \cite{chen-etal-2024-self-para} have been made to reduce SC computational overhead while maintaining quality. However, none have matched the widespread adoption of self-consistency. This can be largely attributed to several limitations: (1) a trade-off where throughput is reduced while \textit{latency increases}, for example by sampling chains sequentially (instead of in parallel) until reaching a certain condition \cite{aggarwal2023let, li2024escape, wang2024make} or running expensive LLM calls instead of the cheap majority voting \cite{yoran2023answering}, (2) the need for manual feature crafting and tuning tailored to each dataset \cite{wan2024dynamic}, (3) promising results on specialized setups \cite{wang2024soft} which did not generalize to standard benchmarks (Table \ref{table:max-ablation}), and (4) as highlighted by \citet{huang2023large}, many of the more sophisticated methods that appear promising actually don't outperform self-consistency when evaluated with a thorough analysis of inference costs. Our approach is different in that CISC adds minimal complexity to self-consistency, and still allows parallel sampling which enables to improve throughput \textit{without compromising latency}, a crucial requirement for many applications.

\paragraph{Self-consistency with confidence.}
Related approaches to CISC's confidence-weighted majority vote were previously explored in both the original self-consistency paper \citet{wang2022self}, that considered a weighted majority using Sequence Probability (\S\ref{sec:metrics}), and in \citet{miao2023selfcheck}, that concluded that verbally \nl{asking the LLM to check its own reasoning is largely ineffective} for improving self-consistency. In both cases, these failures are attributed to the confidence scores being too similar to one another. Our work shows that despite this, the scores contain a useful signal (reflected in the WQD scores; Table \ref{tab:confidence-methods}) that can be utilized by a normalization step prior to aggregation to significantly improve the efficiency of self-consistency. Furthermore, the P(True) method, which achieves the highest WQD scores, has not been previously used for attempting to improve self-consistency.

\section{Discussion}

In this work we introduced CISC, a lightweight extension of self-consistency. Across diverse models, datasets, and confidence extraction methods, CISC consistently outperformed self-consistency, reducing computation costs by over 40\% on average.

The performance gains achieved by using model-derived confidence scores provide a practical evidence that LLMs can effectively judge the quality of their own outputs, contributing to the ongoing debate on this topic \cite{huang2023large, li2024confidence}. This is further strengthened by our qualitative evaluation, revealing significant agreement between model confidence and human assessments of response quality.

Complementing our investigation of LLM self-assessment, we address the crucial aspect of evaluating confidence methods.  Traditional calibration metrics, which assess confidence across different questions, fail to capture a model's ability to distinguish between high and low quality responses to the same question. To overcome this, we introduce the Within-Question Discrimination (WQD) metric and demonstrate its effectiveness.

We encourage future research to explore the integration of model self-confidence into more sophisticated reasoning frameworks like Tree of Thoughts \cite{yao2024tree} or Graph of Thoughts \cite{besta2024graph}, believing that harnessing this inherent capability can further boost performance. Another promising avenue is training models to produce more accurate intrinsic or verbal confidence \cite{lin2022teaching, chaudhry2024finetuning}, which would directly improve CISC and related methods. For instance, recent evidence suggests that a better signal can be derived from the model's internal states, even outperforming P(True) \cite{gekhman2025inside}. Conversely, CISC and WQD can be used to assess the impact of advancements in confidence generation.

\section{Limitations}

\paragraph{Confidence Prompting. } Our confidence extraction prompting approach minimizes the computational overhead (\S\ref{sec:conf-methods}) by using short confidence prompts (less than 5\% of the input and reasoning chain length) that, unlike other works, are appended after the reasoning chain. This allows us to continue to use the auto-regressive cache that was used when the models generated the answer. While this approach is readily implementable within frameworks like HuggingFace \cite{hf}, it may not be universally supported. An alternative one-step prompting approach, which does not rely on prefix caching, is discussed in Appendix \ref{sec:appendix-prompting}.  We opted for the two-step approach in this study to ensure a clear and robust evaluation of CISC, fully mitigating the impact of confidence integration on the generated reasoning paths.

\paragraph{Access to the model's probabilities. } The preferred CISC approach calculates P(True) (as described in \S\ref{sec:conf-methods}) by examining the model's assigned probability to the verbal confidence token.  This method is available in both popular open-weights frameworks (e.g., \citet{hf}) and closed-weights frameworks (e.g., \citet{openaiapi}).  However, this feature may not be universally available across all frameworks.

\paragraph{Human Evaluation. }  The qualitative human evaluation presented in Section \ref{sec:qualitative} provides further support for our claims regarding LLMs' ability to self-assess the correctness of their responses. This evaluation was conducted on the MMLU dataset, which offers a diverse set of single-choice questions.  Extending this analysis to other datasets could offer additional insights.

\paragraph{Additional ablations. } We examined the performance of CISC across several key aspects, focusing on the impact of the choice of confidence extraction method and the impact of the confidence normalization step. Additional ablations could include examining the effect of zero-shot vs few-shot prompting, different choices of normalization techniques, and using trainable confidence methods \cite{lin2022teaching, chaudhry2024finetuning} to improve the performance of CISC.

\section{Ethics Statement}

This work improves LLM reasoning efficiency by introducing a new decoding strategy (CISC). While CISC itself introduces no new ethical issues, LLMs can perpetuate biases and have societal impacts. Responsible LLM development and deployment, including bias mitigation, are crucial. 

\clearpage
\newpage

\bibliography{custom}

\appendix

\clearpage
\appendix

\section{Quantitative example from \S\ref{sec:cisc}}
\label{appendix:example}

Consider a simplified \emph{binary} setting in which there are two possible answers: correct and incorrect.  Given a number of samples $n$ and a probability $p=0.6$ of generating the correct answer, the number of samples with the correct answer follows the Binomial distribution  $X \sim \text{Binomial}(n, p)$. For such distribution, the majority vote is accurate whenever $X > \frac{n}{2}$ and it has $50\%$ chance to be accurate when $X = \frac{n}{2}$ (i.e., a random choice). 

Now, to illustrate how the self-assessment score of LLMs can be helpful, consider that we have an oracle that assigns twice the weight for answers that are correct. In this case, a weighted majority vote would be accurate whenever $X > \frac{n}{3}$ and it has $50\%$ chance to be accurate when $X = \frac{n}{3}$. 

\begin{figure}[h]
\setlength{\belowcaptionskip}{-10pt}
    \centering
    \includegraphics[width=1\linewidth]{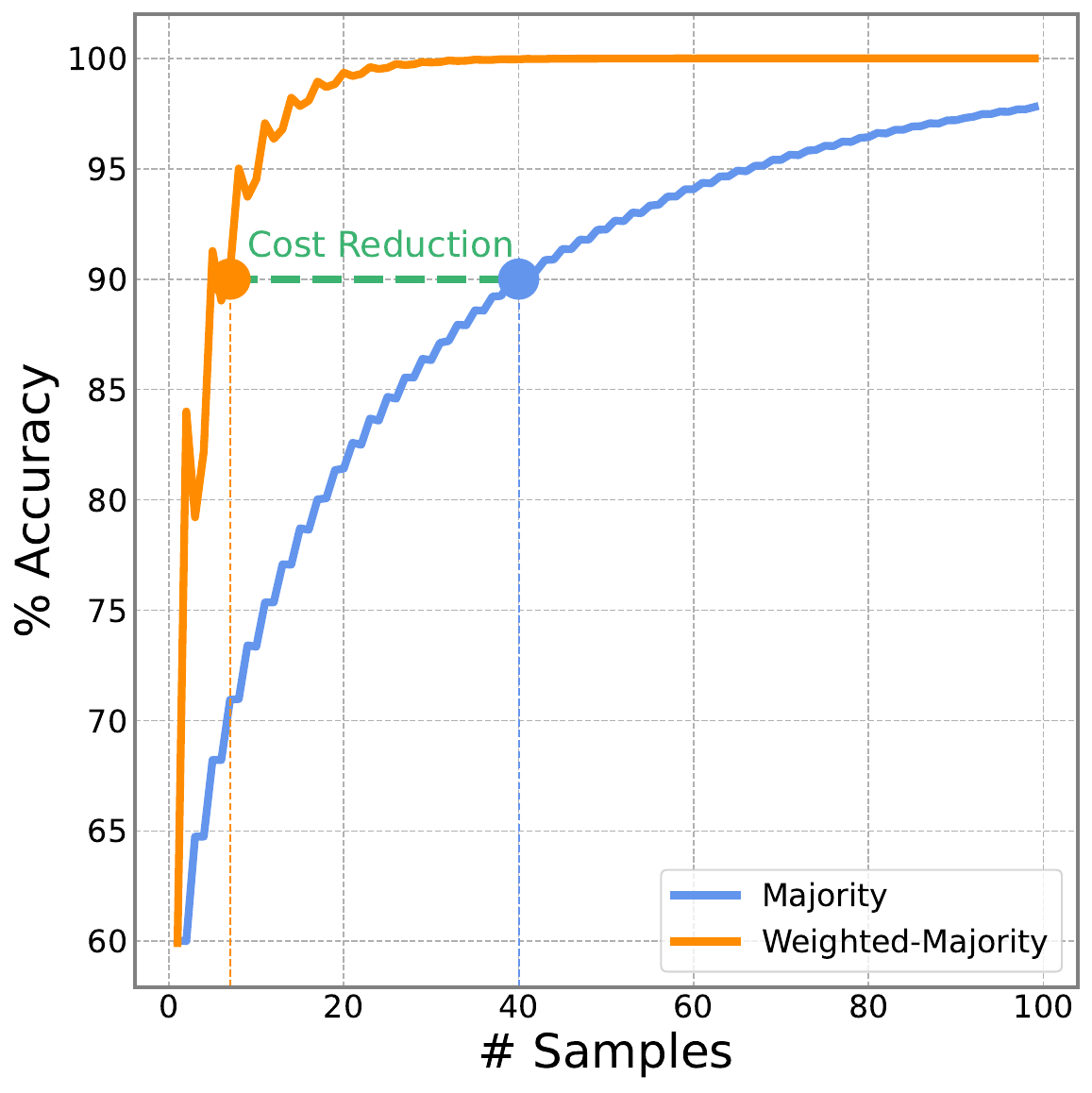}
    \caption{The relationship between the number of samples (x-axis) and the accuracy of majority vote over these samples (y-axis), for two different hypothetical cases sampled from a Binomial distribution: 
     (blue) Each sample receives an equal weight in majority voting, and (orange) Correct answers are assigned double the weight of incorrect ones. Adding this additional weighting information translated into $4X$ reduction in the number of samples required for the majority vote to reach 90\% accuracy.
    }
    \label{fig:thought_experiment}
\end{figure}

In Figure \ref{fig:thought_experiment} we plot the relationship between, (x-axis) the number of samples, and (y-axis) the accuracy of the \emph{weighted} majority vote over these samples. The graph features two lines: (blue) each sample gets an equal weight, and (orange) correct answers are assigned twice the weight of incorrect ones.

While this intuition about cost-saving also applies to the general case of an \emph{arbitrary} set of answers, this setting is trickier to analyze in closed-form  because the specific distribution of incorrect answers impacts the majority vote. E.g., an answer that appears only 20\% of the time can still be correct under majority vote if all the other 80\% incorrect answers are different from one another. This could be obtained by placing additional distributional assumptions on the sampled answers. The analysis of the binary case can be thought of as a worst-case analysis of the general case, since in the worst case, all the incorrect answers are identical and the majority will be accurate if and only if more than half the sampled answers are correct.

\section{Prompting Techniques}
\label{sec:appendix-prompting}

As described in Section \ref{sec:conf-methods}, for our prompt based confidence extraction techniques (Verbal Confidence, P(True)), we used a two-step approach: First, we prompted the model to answer benchmark questions using the prompts shown in Table \ref{tab:question-prompts}. Then, we extracted confidence by concatenating the prompts shown in Table \ref{tab:confidence-prompts} and running the model again. This two-step process allowed using the same answers when comparing self-consistency and CISC.

While a simpler single-step implementation (outputting both answer and confidence in a single response) is possible, we did not explore it in this study. For research purposes, we prioritized a clean setup that ensured requesting confidence scores did not influence the generated answers and chain-of-thoughts.

As shown in Table \ref{tab:confidence-prompts}, all the confidence extraction prompts that we used are extremely lightweight. We deliberately avoided methods that significantly increase the number of generated tokens like generating $k$ guesses with associated probabilities \cite{tian2023just}. 

For the P(True) method, we modified the prompts from \citet{kadavath2022language} in two ways: (1) We changed the format to allow concatenation after the model provided its answer, ensuring that prefix caching could be re-used between the two steps. (2) We changed the prompt format to 0/1 instead of True/False, as some benchmarks are using True/False as ground truth labels, and we observed that it might confuse the model when extracting confidence scores.

\begin{table*}[h]
\centering
\begin{tabular}{c l}
\toprule
\multicolumn{2}{l}{General Instructions} \\ 
\toprule
\multicolumn{2}{l}{\makecell[l{p{15cm}}]{
Before giving your answer, provide a step-by-step explanation of your thought process. Then on a new line, give your proposed answer adhering to this precise format: 'Proposed answer: (X).', where X is your proposed answer.
\\\\
}} \\

\toprule
Dataset  & Prompt \\
\toprule
MMLU-Pro  & \makecell[l{p{11cm}}]{
You will be given a single-choice question. Answer the question by selecting the letter of the
best fitting option.\\\\\emph{[General Instructions]}\\\\The answer MUST ALWAYS
be the letter of one of the available options; it CANNOT be "None of the Above".
} \\
\midrule

MATH      & \makecell[l{p{11cm}}]{
You will be given a question and your goal is to answer it correctly.\textbackslash nYour proposed answer should be a TeX expression, such as '\$5\$', '\$3.14\$', or '\$\textbackslash\textbackslash sqrt\{8\}\$\\\\\emph{[General Instructions]}
} \\
\midrule

\makecell{BBH\\(no options)}      & \makecell[l{p{11cm}}]{
You will be given a question and your goal is to answer it correctly.\\\\\emph{[General Instructions]}
} \\
\midrule
\makecell{BBH\\(with options)}      & \makecell[l{p{11cm}}]{
You will be given a question and your goal is to answer it correctly.\\\\\emph{[General Instructions]}\\\\
Select the letter of the best fitting option. The answer CANNOT be "None of the Above".
} \\

\midrule

GSM8K     & \makecell[l{p{11cm}}]{
You will be given a question and your goal is to answer it correctly.\\\\\emph{[General Instructions]}
} \\
\bottomrule

\end{tabular}
\caption{ The prompts used to generate model responses for benchmark questions. For all datasets, we used the \textit{General Instructions} (shown at the top) asking the model to solve each question step-by-step and provide its final answer in a specified format. In addition, for each dataset we briefly explained the expected questions format. All prompts were zero-shot; few-shot experiments are reserved for future work.
}
\label{tab:question-prompts}
\end{table*}

\begin{table*}[h]
\centering
\begin{tabular}{c l}
\toprule
Confidence Method  & Prompt \\
\toprule
Verbal 0-100  & \makecell[l{p{11cm}}]{
Now I will rate my confidence in the proposed answer on a scale of 0-100.
Proposed confidence: (
} \\
\midrule

Verbal Binary      & \makecell[l{p{11cm}}]{
Now I will rate my confidence in the proposed answer as either 0 or 1.
Proposed confidence: (
} \\
\bottomrule

\end{tabular}
\caption{ The prompts used to extract the model confidence in its response. As explained in section \ref{sec:appendix-prompting}, these prompts are concatenated as a second step, after the model already answers the question. For the P(True) method, we used the Verbal Binary prompt and looked at the probably the model assigns to the token 1. Importantly, in all the models evaluated in this work, "(0" and "(1" are tokenized as two separate tokens. 
}
\label{tab:confidence-prompts}
\end{table*}
\section{Additional Results}
\label{sec:appendix-more-results}

For each confidence method, Table \ref{table:aggregated-results} shows the macro-average results across all models and datasets. A more detailed version of this table, with a per dataset breakdown, is given at Table \ref{tab:apendix-all-results}.

In addition, Table \ref{table:micro-results} shows micro-averaged aggregated results with confidence intervals, demonstrating the strong statistical significance of our findings. These bootstrap confidence intervals were calculated as follows: (1) For each confidence method, results from all datasets and models were combined into a single dataset of approximately $n \approx 150,000$ rows. (2) 10,000 bootstrap sets were generated by repeatedly sampling $n$ elements with replacement. (3) The procedure described in \ref{sec:bootstrap} was applied to each set, yielding 10,000 estimates of the mean accuracy difference. (4) We used these estimates to calculate the 95\% interval.

Table \ref{table:norm-table-ext} is an extended version of table \ref{table:norm-table}. One important insight that can be derived from the extended table, is that using softmax normalization without temperature scaling is strongly discouraged for CISC.

\begin{figure*}[h]
    \centering
    \includegraphics[width=1\linewidth]{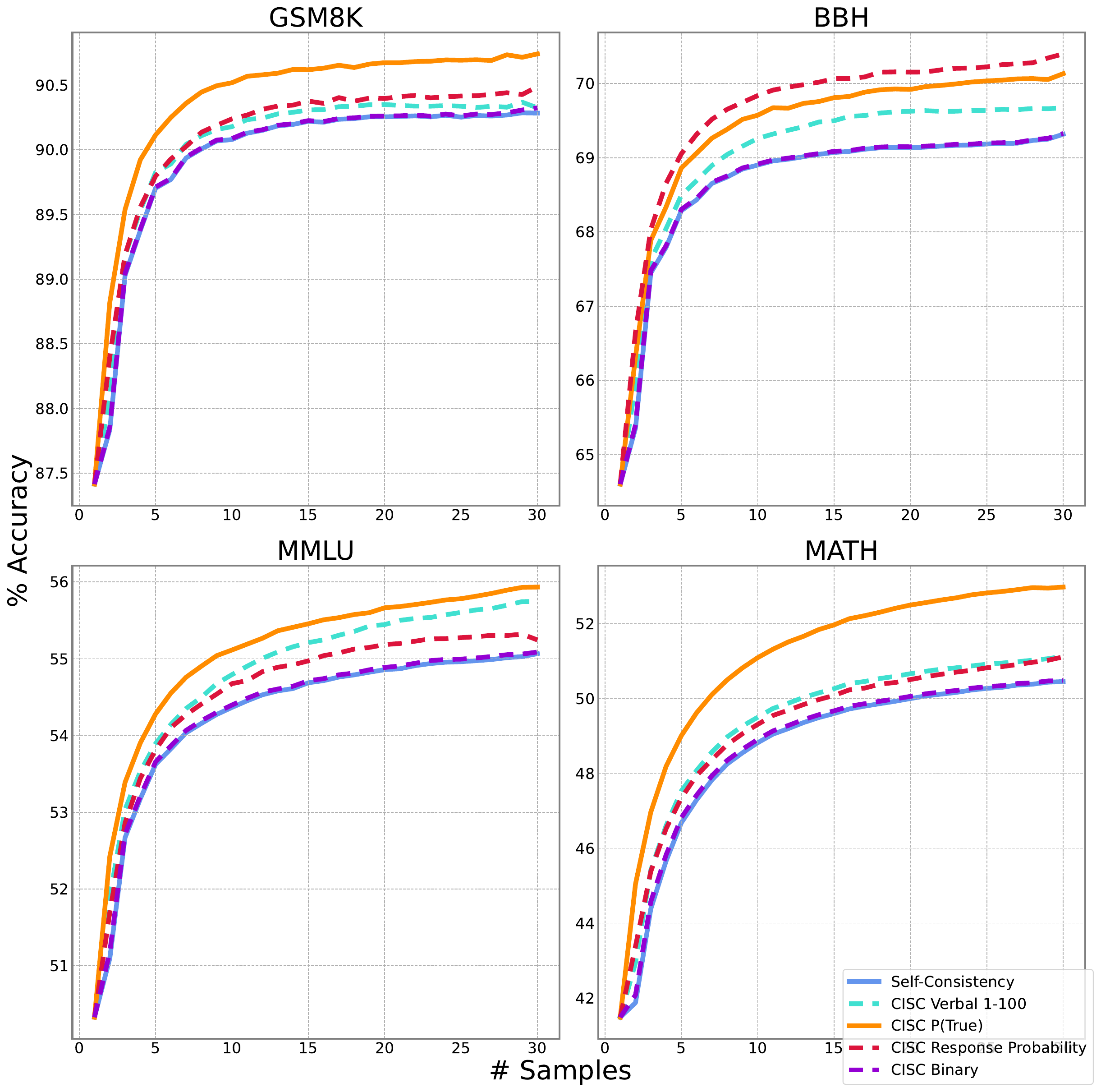}
    \caption{Comparison between different confidence extraction methods using Gemma2-9B model and four datasets (\S\ref{sec:datasets}). CISC with P(True) outperforms Self-Consistency and is the best of all the CISC variants.
    }
    \label{fig:methods-graph}
\end{figure*}

\begin{figure*}[h]
    \centering
    \includegraphics[width=1\linewidth]{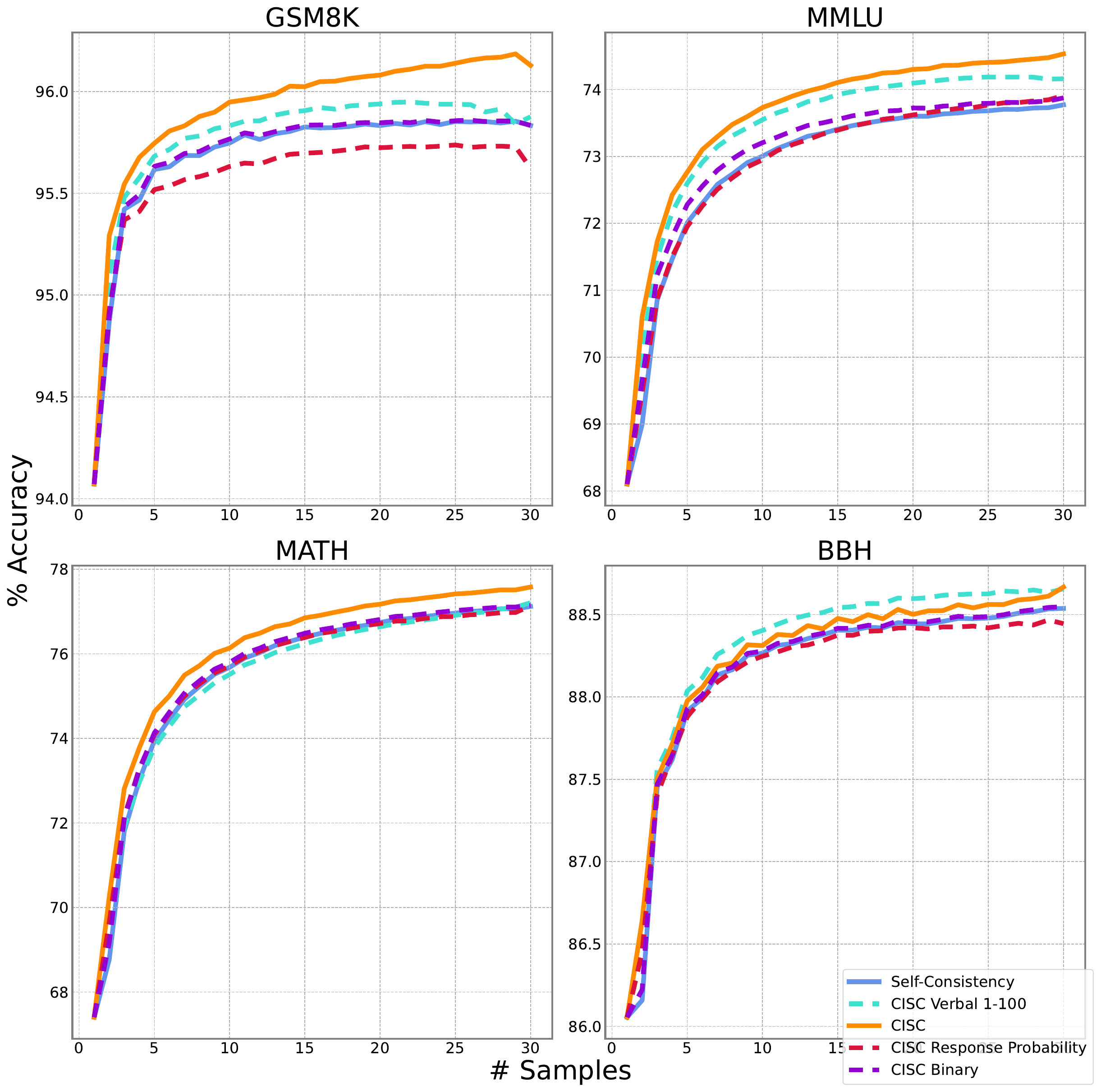}
    \caption{Comparison between different confidence extraction methods using Mistral 123B model and four datasets (\S\ref{sec:datasets}). CISC with P(True) outperforms Self-Consistency in all 4 graphs and is the best of all the CISC variants in 3 graphs.
    }
    \label{fig:methods-graph-mistral123}
\end{figure*}

We also add Figures \ref{fig:methods-graph}, \ref{fig:methods-graph-mistral123} featuring additional graphs similar to Figure \ref{fig:first-figure}, but with all the confidence methods. 

Finally, in Table \ref{table:max-ablation}, we include ablations comparing CISC's weighted majority mechanism to more simple methods like selecting the max confidence \cite{wang2024soft} or using the confidence values as a tie-breaker for self-consistency.

\begin{table*}[!h]
\centering
\begin{tabular}{llllrr}
\toprule
& & \multicolumn{2}{c}{Comparable SC Samples} & \multicolumn{2}{c}{Acc Improvement (\%)} \\
\cmidrule(lr){3-4} \cmidrule(lr){5-6}
Dataset & Confidence Method & Budget 5 & Budget 10 & Budget 5 & Budget 10 \\
\midrule
\multirow[t]{4}{*}{MMLU} & Verbal Binary & 18\% \small{(6.1)} & 12\% \small{(11.3)} & 0.4 & 0.2 \\
 & Verbal 1-100 & 25\% \small{(6.7)} & 32\% \small{(14.6)} & 0.9 & 0.7 \\
 & Response Probability & 17\% \small{(6.0)} & 23\% \small{(13.0)} & 0.7 & 0.6 \\
 & P(True) & \textbf{37\% \small{(7.9)}}& \textbf{47\% \small{(18.8)}} & \textbf{1.4} & \textbf{1.0} \\
\cline{1-6}
\multirow[t]{4}{*}{MATH} & Verbal Binary & 18\% \small{(6.1)} & 11\% \small{(11.2)} & 0.8 & 0.5 \\
 & Verbal 1-100 & 17\% \small{(6.0)} & 12\% \small{(11.3)} & 1.3 & 0.6 \\
 & Response Probability & 19\% \small{(6.2)} & 17\% \small{(12.0)} & 2.2 & 1.2 \\
 & P(True) & \textbf{32\% \small{(7.3)}} & \textbf{34\% \small{(15.2)}} & \textbf{3.0} & \textbf{2.0} \\
\cline{1-6}
\multirow[t]{4}{*}{GSM8K} & Verbal Binary & 18\% \small{(6.1)} & 7\% \small{(10.8)} & 0.2 & 0.1 \\
 & Verbal 1-100 & 22\% \small{(6.4)} & 32\% \small{(14.6)} & 0.3 & 0.1 \\
 & Response Probability & 21\% \small{(6.3)} & 33\% \small{(14.9)} & 0.7 & 0.5 \\
 & P(True) & \textbf{43\% \small{(8.8)}} & \textbf{53\% \small{(21.2)}} & \textbf{0.9} & \textbf{0.6} \\
\cline{1-6}
\multirow[t]{4}{*}{BBH} & Verbal Binary & 17\% \small{(6.0)} & 10\% \small{(11.1)} & 0.2 & 0.1 \\
 & Verbal 1-100 & 22\% \small{(6.4)} & 41\% \small{(17.0)} & 0.5 & 0.4 \\
 & Response Probability & 32\% \small{(7.3)} & 45\% \small{(18.3)} & 0.7 & 0.8 \\
 & P(True) & \textbf{48\% \small{(9.7)}} & \textbf{47\% \small{(19.0)}} & \textbf{1.0} & \textbf{0.9}\\
\bottomrule
\end{tabular}
\caption{ Aggregated results across all models for each dataset and confidence extraction method.  All methods demonstrate better performance than standard self-consistency, with the P-True method achieving the best results and leading to an computational cost reduction of up to 53\% }
\label{tab:apendix-all-results}
\end{table*}

\begin{table*}[!ht]
\centering
\begin{tabular}{lcccc}
\toprule
& \multicolumn{2}{c}{Acc Improvement} \\
\cmidrule(lr){2-3} \cmidrule(l){4-5} 
Confidence Method & Budget 5 & Budget 10 \\
\midrule
Verbal Binary & 0.35\ \small{(0.34-0.37)} & 0.20\ \small{(0.18-0.21)} \\[4pt]
Verbal 1-100 & 0.68\ \small{(0.64-0.72)} & 0.46\ \small{(0.40-0.51)} \\[4pt]
Response Probability & 0.88\ \small{(0.84-0.92)} & 0.69\ \small{(0.63-0.74)} \\[4pt]
P(True) & \textbf{1.38\ \small{(1.32-1.43)}} & \textbf{1.03\ \small{(0.96-1.10)}} \\[4pt]

\bottomrule
\end{tabular}
\caption{\textbf{Micro-averaged Aggregated Results. } This table presents the micro-averaged aggregated results with confidence intervals for each confidence method. Each confidence method demonstrates statistically significant improvements over self-consistency, and \textbf{P(True)} method exhibits significant superiority over other methods. This detailed view supplements the macro-average results shown in Table \ref{table:aggregated-results} and provides statistical verification of the efficiency gains and accuracy improvements attributed to CISC methods. }
\label{table:micro-results}
\end{table*}

\begin{table*}[!h]
\centering
\begin{tabular}{llllll}
\toprule
 & \multicolumn{2}{l}{\% Cost Reduction} & \multicolumn{3}{l}{\% Acc Improvement} \\
 & 5 & 10 & 5 & 10 & 30 \\
Confidence Method &  &  &  &  &  \\
\midrule
P(True) - No Normalization & 29\% \small{(7.0)} & 32\% \small{(14.8)} & 1.4 & 0.8 & 0.4 \\
P(True) - Softmax T=1 & 27\% \small{(6.8)} & 30\% \small{(14.2)} & 1.3 & 0.8 & 0.3 \\
P(True) - Softmax T=Tuned & \textbf{41\% \small{(8.4)}} & \textbf{46\% \small{(18.6)}} & \textbf{1.6} & \textbf{1.1} & \textbf{0.9} \\
\midrule
Sequence Probability - No Normalization & 21\% \small{(6.3)} & 24\% \small{(13.1)} & 1.1 & 0.6 & 0.3 \\
Sequence Probability - Softmax T=1 & 20\% \small{(6.3)} & 23\% \small{(13.0)} & 1.1 & 0.6 & 0.2 \\
Sequence Probability - Softmax T=Tuned & \textbf{22\% \small{(6.5)}} & \textbf{31\% \small{(14.6)}} & \textbf{1.1} & \textbf{0.8} & \textbf{0.7} \\
\midrule
Verbal 0 - 100 - No Normalization & 20\% \small{(6.3)} & 20\% \small{(12.5)} & 0.7 & 0.4 & 0.1 \\
Verbal 0 - 100 - Softmax T=1 & 12\% \small{(5.7)} & -1\% \small{(9.9)} & -0.3 & -1.4 & -2.6 \\
Verbal 0 - 100 - Softmax T=Tuned & \textbf{22\% \small{(6.4)}} & \textbf{30\% \small{(14.4)}} & \textbf{0.8} & \textbf{0.4} & \textbf{0.3} \\
\bottomrule
\end{tabular}
\caption{\textbf{Normalization Ablation. } This table extends Table \ref{table:norm-table}, showing that temperature-scaled softmax is optimal for all methods, and that softmax should be avoided without temperature scaling. }
\label{table:norm-table-ext}
\end{table*}

\begin{table*}[!ht]
\centering
\begin{tabular}{lcccc}
\toprule
& \multicolumn{2}{c}{Comparable SC Samples} & \multicolumn{2}{c}{Acc Improvement (\%)} \\
\cmidrule(lr){2-3} \cmidrule(lr){4-5}
Confidence Method & Budget 5 & Budget 10 & Budget 5 & Budget 10 \\
\midrule
Max & -11\% \small{(4.5)} & -84\% \small{(5.4)} & -1.9 & -4.5 \\[4pt]
Tie & 27\% \small{(6.8)} & 28\% \small{(13.9)} & 1.3 & 0.7 \\[4pt]
CISC & \textbf{41\% \small{(8.4)}} & \textbf{46\% \small{(18.6)}} & \textbf{1.6} & \textbf{1.1} \\[4pt]

\bottomrule
\end{tabular}
\caption{\textbf{Simplified ablation. } Here we compare CISC with two simplified ablations:  (Max) Which selects the answer with highest confidence score, and (Tie) Only uses CISC if self-consistency resulted in a tie. All methods are calculated using the P(True) confidence. Results are aggregated across all models and datasets. CISC significantly outperforms both ablations, and the Max method even degenerates performance.}
\label{table:max-ablation}
\end{table*}

\begin{table*}[!h]
\centering
\begin{tabular}{lcccc}
\toprule
Dataset & BBH & GSM8K & MATH & MMLU \\
Model &  &  &  &  \\
\midrule
Gemma 27b & 57.1 & 66.1 & 62.9 & 59.9 \\
Gemma 2b & 55.8 & 66.2 & 64.3 & 53.6 \\
Gemma 9b & 55.3 & 68.3 & 71.8 & 58.9 \\
Mistral 123 & 56.2 & 66.1 & 61.2 & 63.4 \\
Mistral 22 & 64.1 & 81.4 & 74.9 & 67.7 \\
Mistral 8 & 59.4 & 71.8 & 62.9 & 58.8 \\
Qwen 14b & 58.9 & 65.5 & 59.0 & 60.2 \\
Qwen 3b & 56.3 & 61.9 & 57.5 & 56.0 \\
Qwen 72b & 53.5 & 62.4 & 63.6 & 58.8 \\
\bottomrule
\end{tabular}
\label{table:wqd-breakdown}
\caption{\textbf{Within-Question-Discrimination Breakdown.} This table presents a breakdown of the aggregated Within-Question-Discrimination (WQD) results presented in Table \ref{tab:confidence-methods}, using the P(True) method.  In all cases, WQD scores exceed the 50\% chance level. 
}
\end{table*}
\section{Temperature Scaling Results}
\label{sec:appendix-temperature}

\begin{figure}[h]
    \centering
    \includegraphics[width=0.9\linewidth]{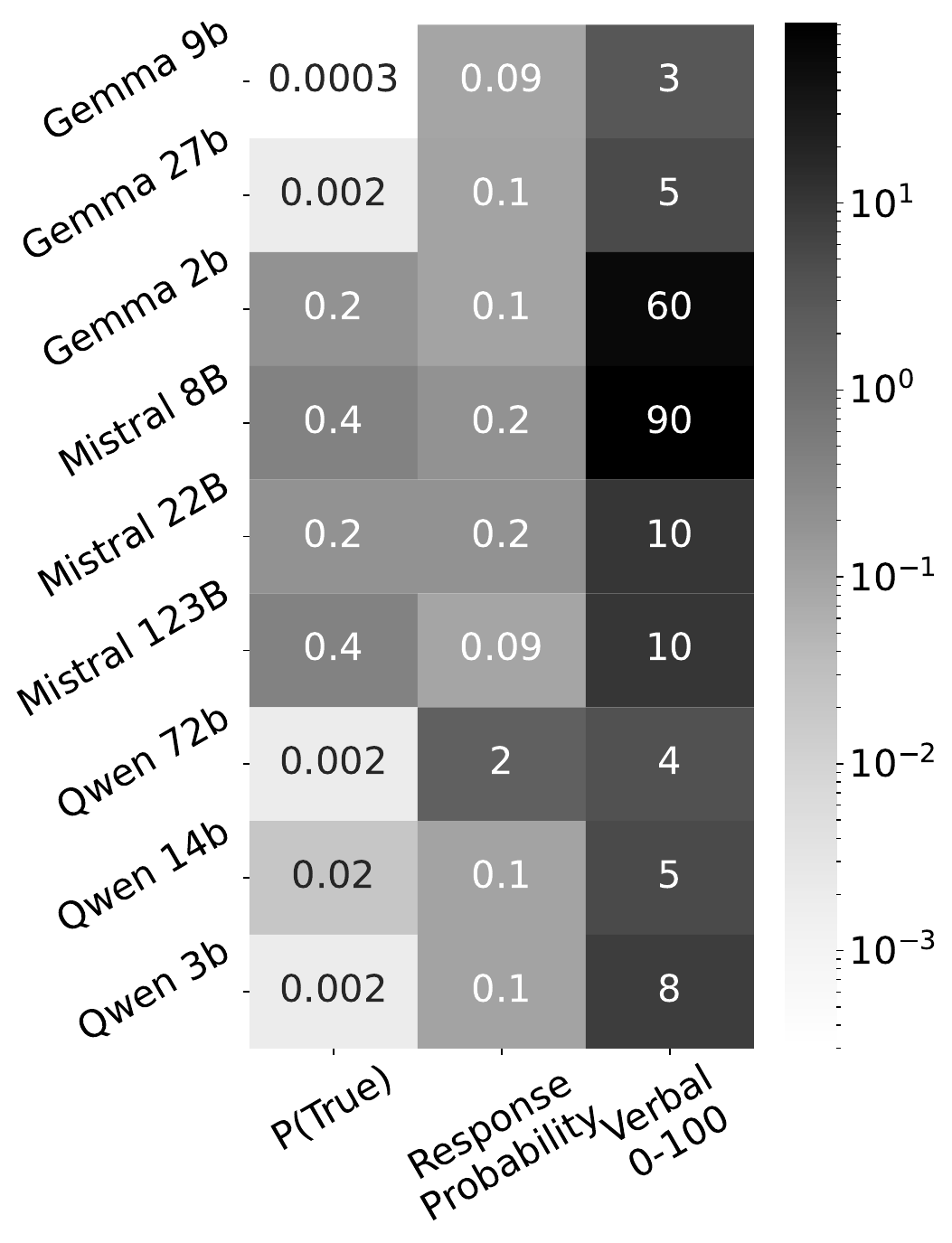}
    \caption{The best temperatures values for each model / confidence-method combination. As discussed in Section \ref{sec:temperature}, we fit a single temperature hyper-parameter across 10\% of all datasets together. As can be seen, each of the confidence extraction method work with a different temperature magnitude. We also see variability between models using the same confidence extraction method.
    }
    \label{fig:temperatures-map}
\end{figure}

As discussed in \S\ref{sec:temperature}, a single optimal temperature, $T^\star$, was determined for each model and confidence extraction method by using a 10\% held-out set, aggregated across all datasets. Fitting is done using grid search on 80 evenly spaced values ranging from $10^{-4}$ to $10^{4}$. This was a light-weight process, only taking a few minutes on a standard desktop since no LLM re-runs were necessary. The temperatures for each configuration are presented in Figure \ref{fig:temperatures-map}. As can be seen, each of the confidence extraction method work with a different temperature magnitude because it produce confidence values on a different scale.

\section{Qualitative Appendix}
\label{sec:appendix-qualitative}

The qualitative analysis presented in \S\ref{sec:qualitative} involved sampling the reasoning paths using three models: Qwen2.5 3B, Gemma2 9B and Mistral Large (123B). To broaden our evaluated sample pool, we employed a bootstrap process, sampling three distinct traces per question multiple times. Then, we first filtered these samples so that each of them arrived from a different question, and continued with the sampling process described in \S\ref{sec:qualitative}.  

Human evaluators were asked to identify logical patterns in the LLMs' reasoning paths that reduced the evaluators' confidence in the correctness of the LLMs' answers. Importantly, the MMLU dataset requires significant domain knowledge and unspecialized humans achieved only 34.5\% accuracy \cite{hendrycks2020measuring}, compared to a random baseline of 25\%. The MMLU-pro dataset is based on the MMLU dataset, but is considered much harder. This means that our evaluators, which lacked specialized knowledge, could not easily how to solve each question. Instead, we instructed them to focus on identifying low-quality reasoning errors in the responses of the LLMs. This approach aligns with findings from a prior analysis on GPT-4o \cite{wang2024mmlupro}, which attributed 39\% of its errors to reasoning flaws that do not rely on specialized domain knowledge.

Following this review, we aggregated the indicators of low quality into high-level categories. Three main categories encompassed 49\% of the samples. The remaining samples either lacked low-quality indicators (50\%) or had indicators that did not fit into a sizable category (1\%). The different categories and their prevalence are presented in Table \ref{table:patterns}. 

Two of these three categories show a strong association with relative-low confidence scores from the model: (1) The model arrived at solutions not present among the available options, and (2) The model only conducted partial calculations necessary. Interestingly, the pattern where the model explores several plausible solutions without identifying a definitive "correct" one was not specifically associated with either high or low confidence in the model’s reasoning paths, underscoring that not all human-identified patterns significantly influence the model’s assessment. 

Overall, the alignment of human-identified low-quality indicators with low-confidence scores provides another evidence of the ability of LLMs to self-assess and prioritize high confidence solutions. An ability that is leveraged by CISC.

\begin{table*}[!h]
\centering
\begin{tabular}{lllll}
\toprule
\multicolumn{1}{c}{\textbf{Category}} & \multicolumn{1}{c}{\textbf{Definition}} & \multicolumn{1}{c}{\textbf{Low}} & \multicolumn{1}{c}{\textbf{High}} & \multicolumn{1}{c}{\textbf{Snippet}} \\
\midrule
No choice &
\makecell[l{p{5cm}}]{
The model arrives at a solution which is not present in the list of available options. This can include case where a mathematical answer significantly diverging from all options, answers that are only partially correct, or the elimination of all options as part of the reasoning process.
} & 38\% & 13\% & 
\makecell[l{p{5cm}}]{"... After reviewing the options, it's clear that none of 
them perfectly fit the requirements. 
However, the closest correct option is (A), which only has 
a minor error in calculating the remaining inches. 
Proposed answer: (A)"} \\
\midrule
\makecell[l]{Incomplete \\ Calculations} & 
\makecell[l{p{5cm}}]{
The model begins to solve the problem but does not complete the full calculation, often due to the lack of necessary data. 
For example, when attempting to compute acceleration, the absence of mass data prevents an exact and full calculation.
} & 22\% & 2\% & 
\makecell[l{p{5cm}}]{"...**Calculate Heat Flow:** q" = h * (T-surface - T-air) 
**Note:**  Without the actual values for air density, viscosity, 
and thermal conductivity at 68°F, we cannot perform  
the precise calculations. 
Proposed answer: (C)." } \\
\midrule
\makecell[l]{Multiple \\ candidates} & 
\makecell[l{p{5cm}}]{
The model explores several plausible solutions without 
identifying a definitive "correct" one. This occurs when 
the model solves a problem generally, relying on 
estimations rather than concrete data, resulting in a 
range of potential answers.
} & 11\% & 16\% & 
\makecell[l{p{5cm}}]{"... 2.**Identify Buddhist Thinkers:** The options list several 
prominent Buddhist figures from various traditions...
4. **Most Prominent:** The Dalai Lama and Thich Nhat 
Hanh stand out for their consistent emphasis 
on self-sacrifice in their teachings and actions. 
Proposed answer: (I)"} \\
\bottomrule
\end{tabular}
\caption{Human evaluators identified low-quality reasoning indicators in LLM responses (see \S\ref{sec:qualitative}). These indicators were then clustered into three categories, each described above with a definition and an example snippet from an LLM response. The (Low, High) columns show the percentage of LLM responses with low/high self-assessed confidence that exhibited each pattern.  The "No Choice" and "Incomplete Calculation" categories are strongly associated with low confidence.}
\label{table:patterns}
\end{table*}

\begin{table*}[!h]
\centering
\begin{tabular}{ll}
\\
\toprule
\\
\makecell[l]{\textbf{General} \\  \textbf{Instructions}}
& \makecell[l{p{12cm}}]{ Evaluate the LLMs' reasoning paths, looking for logical inconsistencies or errors that lower your confidence in their conclusions.  Because the questions are very difficult, even for experts, your task is to identify general reasoning flaws, not to assess the correctness of the final answers themselves. 
Examples: 
\begin{itemize}
    \item Incorrect Assumption: The model assumes something without justification
    \item Missing Step: The model skips a crucial step in the reasoning process
    \item Contradiction: The model states both A and not-A
\end{itemize}
} 
\\
\hline
\\
\textbf{Question}
& \texttt{[Pre-filled - The original question given to the LLM]} \\
\\
\hline
\\
\makecell[l]{\textbf{LLM} \\  \textbf{Output}} &
\texttt{[Pre-filled - The LLM output for the given question]} \\
\\
\bottomrule
\\
\end{tabular}
\caption{The input given to human evaluators as part of our qualitative analysis (\S\ref{sec:qualitative}).}
\label{tab:evaluation_template}
\end{table*}

\section{Compute}
\label{sec:appendix-compute}

For each model (\S\ref{sec:models}), we generated approximately 500,000 responses - 17,000 questions (\S\ref{sec:datasets}), with 30 samples (\S\ref{sec:bootstrap}). As a reference, inference with Gemma2-2-Billion (1K token context length) required an order of 100 Nvidia H100 GPU hours.

We consider 30 samples to be a substantial sample size. In more practical scenarios, we anticipate practitioners would likely use a smaller number of samples. As illustrated in Figures \ref{fig:first-figure} and \ref{fig:methods-graph}, the improvement curves show a logarithmic shape. On average, across all models and datasets, we found that just 13 responses were sufficient to achieve 90\% of the maximum effect observed with 30 responses.

\end{document}